\documentclass[manuscript]{acmart}

\usepackage{subcaption}

\usepackage{soul}
\usepackage{siunitx}
\usepackage{comment}
\usepackage{todonotes}
\usepackage{amsfonts}
\usepackage{graphicx}
\usepackage{float}

\AtBeginDocument{%
  \providecommand\BibTeX{{%
    \normalfont B\kern-0.5em{\scshape i\kern-0.25em b}\kern-0.8em\TeX}}}

\setcopyright{acmcopyright}
\copyrightyear{2021}
\acmYear{2021}
\acmDOI{}

\begin{document}

\title{Learning Nigerian accent embeddings from speech: preliminary results based on SautiDB-Naija corpus.}

\author{Tejumade Afonja}
\email{teju.afonja@aisaturdayslagos.com}
\affiliation{%
  \institution{AI Saturdays Lagos}
  \state{Lagos}
  \country{Nigeria}
}

\author{Oladimeji Mudele}
\affiliation{%
  \institution{Independent researcher}
  \streetaddress{}
  \city{Pavia}
  \state{Lombardy}
  \country{Italy}
  \postcode{27100}}
\email{mudeledimeji@gmail.com}

\author{Iroro Orife}
\affiliation{%
  \institution{Niger-Volta Language Technologies Institute}
   \city{Seattle}
  \state{Washington}
  \country{USA}}
\email{iroro@alumni.cmu.edu}

\author{Kenechi Dukor}
\affiliation{%
  \institution{AI Saturdays Lagos}
  \streetaddress{}
  \state{Lagos}
  \country{Nigeria}
  \postcode{}}
\email{kenechi.dukor@aisaturdayslagos.com}

\author{Lawrence Francis}
\affiliation{%
 \institution{AI Saturdays Lagos}
  \streetaddress{}
  \city{Lagos}
  \country{Nigeria}}
\email{lawrencedikeu@gmail.com}

\author{Duru Goodness}
\affiliation{%
  \institution{AI Saturdays Lagos}
  \streetaddress{}
  \city{Lagos}
  \country{Nigeria}}
\email{goodyduru@gmail.com}

\author{Oluwafemi Azeez}
\affiliation{%
  \institution{AI Saturdays Lagos}
  \city{Lagos}
  \country{Nigeria}}
\email{azeezfemi17937@yahoo.com}

\author{Ademola Malomo}
\affiliation{%
  \institution{AI Saturdays Lagos}
  \state{Lagos}
  \country{Nigeria}}
  \email{demlabz@gmail.com}
  
\author{Clinton Mbataku}
\affiliation{%
  \institution{AI Saturdays Lagos}
  \city{Lagos}
  \country{Nigeria}}
\email{mbatakuclinton@gmail.com}



\renewcommand{\shortauthors}{Afonja, et al.}


\begin{abstract}
This paper describes foundational efforts with SautiDB-Naija, a novel corpus of non-native (L2) Nigerian English speech. We describe how the corpus was created and curated as well as preliminary experiments with accent classification and learning Nigerian accent embeddings. The initial version of the corpus includes over 900 recordings from L2 English speakers of Nigerian languages, such as Yor{\`{u}}b{\'{a}}, \`{I}gb\`{o}, \d{\`{E}}d\'{o}, Efik-Ibibio, and Igala. We further demonstrate how fine-tuning on a pre-trained model like wav2vec can yield representations suitable for related speech tasks such as accent classification. SautiDB-Naija has been published to Zenodo for general use under a flexible Creative Commons License.
\end{abstract}




\maketitle

\section{Introduction}
Advances in education, technology and transportation have made the world a much smaller place, with many of us from different cities, regions and countries commonly speaking a hypercentral or global language~\citep{de2013words}. In recent years, online courses and educational resources have become widely available, most of which are offered in English ~\citep{mamgain2014learner, vinke1998english}. These courses are intended to be accessible to learners from around the world, yet students must often adapt to the teaching style of the instructor, delivering the course in an unfamiliar accent. This can place a significant cognitive burden on students and put them at a disadvantage vis-\`{a}-vis peers who are familiar with the instructor's accent. It would therefore be desirable to present online video content to learners in a familiar accent. Motivated by the potential of accent transfer technology to facilitate better learning experiences, our contributions are as follows:

\begin{enumerate}
    \item {\bf SautiDB-Naija \citep{afonja_tejumade_2021_4561842}}, an accented speech corpus of non-native English speech consisting of 919 speech recordings from an assemblage of first-language (L1) speakers of over five Nigerian languages, Yor{\`{u}}b{\'{a}}, \`{I}gb\`{o}, \d{\`{E}}d\'{o}, Efik-Ibibio, and Igala. 
    \item {\bf SautiClassify}, a first attempt to quantify the accent information axis of the SautiDB-Naija corpus. In other words, we address the question: Can the information about the different accents in the SautiDB-Naija corpus be characterised by a neural network model? To answer this question, we designed an experiment to learn embeddings of the different accents in the corpus. In the following sections, we present details of our experiments and results.
\end{enumerate}

\section{Related Works}

The are a number of existing or accomplished speech data collection projects but most of these works do not focus on accented English data~\citep{kominek2004cmu}. 

In Ref.~\citep{zhao2018l2}, the L2-Arctic speech corpus has been presented. This mentioned corpus  includes recordings of speakers whose first languages are Hindi, Korean, Mandarin, Spanish, and Arabic. While the  L2-Arctic speech corpus extends across broad categories of accents in different continents of the world, it misses subtle details within sub-regions and ethnic groups. 

The authors of~\citep{ahamad2020accentdb} presented AccentDB: a collection of five different Indian-English accents. In addition, that same study presented and compared different neural networks-based (multi-layer perceptron and 1D convolutional neural network) accent classification models. Ref. \citep{ardila2019common} also provides an accent corpus at sub-national scale. While this corpus contains about 1\% Southern African accents, it doesn't contain any Nigerian accent.

Google Nigerian English~\citep{google_nigerian_english} is a high quality speech corpus of collected over a wide range of Nigerian English speakers and with over 3000 spoken sentences. The speech samples of that corpus are, however, not annotated with accent information. This makes it unusable for speech accent classification or recognition tasks.

For audio classification tasks, the input features are of prime importance. As a result, choosing the appropriate feature representation for chosen tasks remains a growing research area. The study presented in~\citep{perez2019cnn} analysed different handcrafted log-Mel representations and combinations for acoustic signals classification. The result of that study asserts the need careful experimentation in choosing classification task input features.  

Along similar lines, Ref.~\citep{behravan2014introducing} proposed a hybrid approach for foreign accent recognition by combining both phonotactic and spectral-based features. Specifically, the method in that study involves the extraction of speech attribute features and acoustic cues reflecting foreign accents of a speaker to obtain feature streams that are modeled with the i-vector~\citep{ibrahim2018vector} methodology. The authors of Ref.~\citep{zhang2021accent} obtained hybrid phonetic features aggregated from jointly-trained acoustic speech recognition  and accent recognition tasks to achieve robust accent recognition. The logarithmic Mel-filter bank coefficients (FBANK) feature representation was used as modeling input in that study.

Although research in this area has shown that cramming different handcrafted features into neural networks might be sufficient to classify or identify speaker accent, the task of selecting the right input feature to apply remains non-trivial. As a result, it suffices to consider the use of a  pretrained neural network-based acoustic representation as input feature. To this end, our study considers wav2vec~\citep{schneider2019wav2vec}: a model trained with a contrastive loss on large amounts of unlabeled audio data and with representations that improve several downstream classification tasks. 

On the data side of things, there is the need for more accented speech corpus that narrow deeply into chosen parts of the world. We propose SautiDB-Naija as one of such corpora.

\section{Corpus Creation}
\subsection{Crowdsourcing speech recordings}
Firstly, text prompts (sentences) were created from 1132 phonetically balanced sentences in the CMU Arctic Database~\citep{kominek2004cmu}. The text is non-proprietary and the Arctic corpus itself has proven effective in various tasks including speech synthesis~\citep{zen2007hmm}, speech conversion~\citep{zhao2019foreign}, and accent conversion~\citep{ahamad-anand-bhargava:2020:LREC}. Next, a simple web application \cite{sautidb2020}  was designed, implemented and deployed for data collection using Angular and Firebase~\citep{jain2014angularjs, firebase}. An audio API was integrated into the platform for flexible collection of speech samples from speakers with different devices. The app presents various randomly selected sentences to the speakers and prompts them to record their voice while reading the text. The speech recordings are persisted as mono, 16bit, \SI{48}{\kilo\hertz} files.

\subsection{Curation and post-processing}
\begin{figure}[h] 
    \centering
    \includegraphics[width=.4\textwidth]{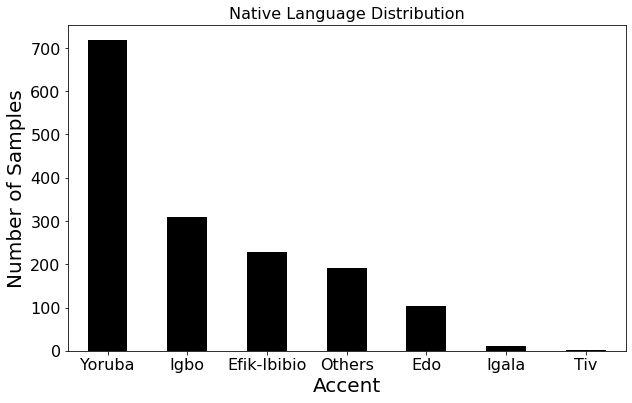} 
  \caption{SautiDB-Naija accent distribution.}
  \label{f:nativelang}
\end{figure}

To prepare the corpus for use, post-processing tasks were performed on the audio recordings to remove accidental repetitions, noisy, muffled or distorted speech and invented sentences not contained in the list of prompts. Next, gender (binary) and sentence IDs were assigned manually to each sample. A verification task is then set up to check for agreement between assigned labels and true labels, and corrections are made in cases of wrong labelling. Finally, the data is normalized to -0.1dB, and all the leading and trailing silences are removed
\footnote{https://github.com/AISaturdaysLagos/sautidb\_postprocessing\_scripts}
. Files were further standardized for upload to the Zenodo Open Repository \cite{afonja_tejumade_2021_4561842}. Figure ~\ref{f:nativelang} shows the distribution of the accents present in the corpus.  

\section{Modeling}
\begin{figure}[h] 
    \centering
    \includegraphics[width=.70\textwidth]{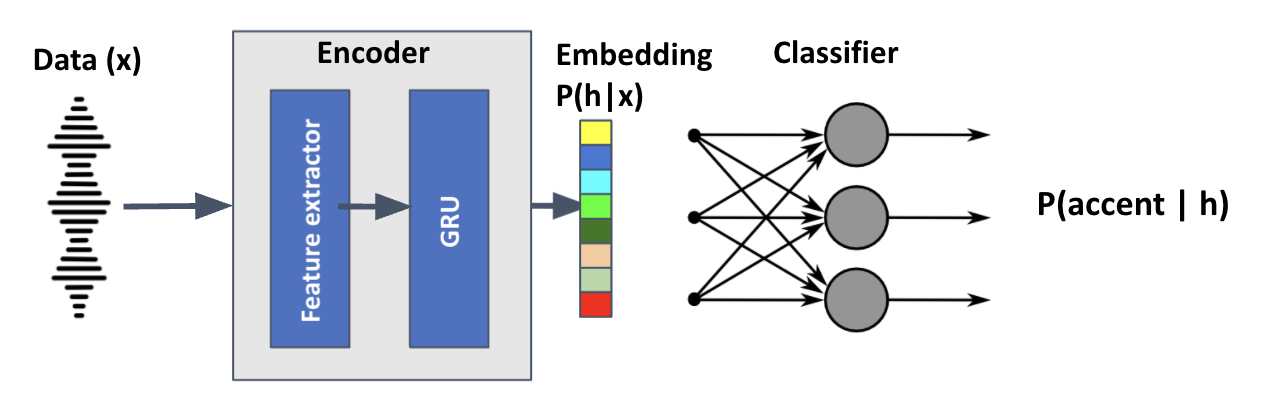} 
  \caption{Our model architecture.}
  \label{f:architecture}
\end{figure}

To demonstrate the fidelity of the accent information contained in the SautiDB-Naija corpus, an accent classification task using an encoder-classifier model was developed. The encoder comprises a feature extractor and a single-layer gated recurrent unit (GRU) recurrent neural network. The feature extractor is applied to the raw audio, $\textbf{x} \in \mathbb{R}^t$ to map into a latent feature $\textbf{z} \in \mathbb{R}^u$, with $z_i \in \mathbb{R}^n$ being the $i$-th element of sequence $\textbf{z}$, and $t >> u$. While $t$ is the temporal size of the input audio, $u$ and $n$ are the temporal size and channel dimension of the latent representation feature, respectively. The GRU is then applied on $\textbf{z}$ to aggregate its sequential information into a single time-independent embedding vector $h \in \mathbb{R}^m$, $m$ being the learned embedding size. A linear layer is then applied to $h$ to predict the accent of the speaker in the input audio. A schematic of the described architecture is presented in Fig.~\ref{f:architecture}.

\subsection{Features extractor}
This forms a critical component of the encoder. Methods for speech processing commonly operate on a variety of handcrafted input features. Such features include the Mel-frequency Cepstral Coefficients (MFCCs) or mel-spectrograms which are a frequency domain representations of speech signals~\citep{melspec}. As an alternative to these hand-crafted features, neural networks-based representation learning models have been shown to produce more robust and generalizable features~\citep{bengio2013representation}. Wav2vec~\citep{schneider2019wav2vec} is trained using a contrastive loss function in self-supervised way on large amounts of unlabeled audio data, resulting in latent representations that generalize into many tasks. 

Our study applies a pretrained wav2vec encoder as feature extractor and compares it to the Mel-spectrogram (80 channels) baseline. We obtain the wav2vec pretrained model using the Wavencoder \texttt{Python} library~\citep{wavencoder}. The wav2vec model is then fine-tuned alongside the remaining components of the model. In addition, we examined the effect of batch normalization of the wav2vec features on the model performance.

\subsection{Training}
We made corpus splits in such a way that all audio files with the same speaker ID are only present in one data split. This allows for a fair evaluation of accent generalization. As a result, we retained only \d{\`{E}}d\'{o}, Yor{\`{u}}b{\'{a}} and \`{I}gb\`{o} accents which have enough unique speaker IDs and thus performed only classification for these accent classes. The splits were obtained as 80\% training, 10\% development, and 10\% test. 


We use a batch-size of 32 for both training and development set, with a random chunk of 3 seconds obtained for each sample per training epoch. A learning rate of 0.01 is applied with the ADAM optimizer. The cross entropy loss~\citep{zhang2018generalized} is used as the objective function to optimize the model. Each fitted model is trained for 50 epochs and the best model (with lowest loss) on the development set is saved.

\section{Results}

Since this study only presents preliminary results, the objective for the experiments presented here is to show that it is possible, using a model, to differentiate different accents in speech utterances in SautiDB-Naija corpus. We achieve this objective by comparing the performances of our trained models to that of the base model with pretrained wav2vec encoder weights as features extractor and random weights (seed = 42) in the GRU and linear classifier head. Our results show promise on the fidelity of accent information in our corpus. In particular, we obtain our best model using a combination of wav2vec and batch normalization. These results are presented in in Table~\ref{t:quality_measures}.

\begin{figure}[h]
     \centering
         \centering
         \includegraphics[width=0.4\textwidth]{"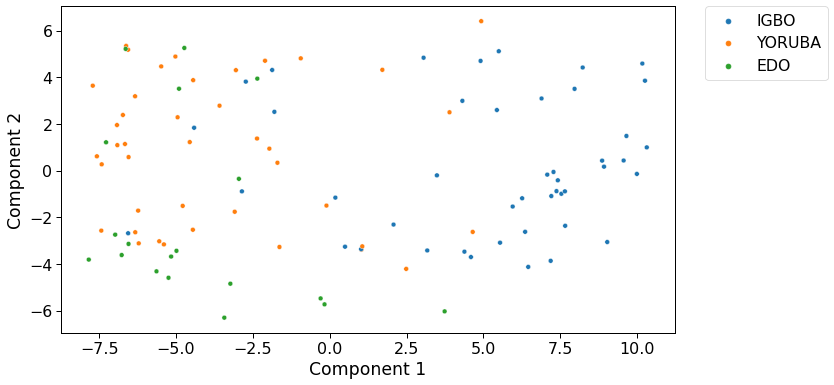"}
         \label{fig:y equals x}
        \caption{Principal Component Analysis~\citep{pca} projection of the embeddings from our best model. While there are some overlaps, it can be seen that accents tend to form clusters in the embedding space.}
        \label{f:scatterplots_of_pca}
\end{figure}

Furthermore, Fig.~\ref{f:scatterplots_of_pca} presents a two-dimensional projection of the accent embeddings obtained with our best model on the test dataset. It can be seen that the embedding space is partitioned by different accents. Furthermore, as presented, Yor{\`{u}}b{\'{a}} and \d{\`{E}}d\'{o} accents which derive from historically overlapping ethnic groups are closer together and both far away from the \`{I}gb\`{o} accent cluster.

\begin{table}[h]
    \centering
\begin{tabular}{ lcc } 
 \textbf{Model} & \textbf{Accuracy} & \textbf{F1-score}\\
 \toprule
 Base model & 0.2667 & 0.2413\\ 
 Mel+GRU & 0.4140 & 0.3809\\ 
 wav2vec+GRU & 0.5333 & 0.4881\\ 
  wav2vec+GRU+BN & \textbf{0.6952} & \textbf{0.6457}\\ 
 \bottomrule
\end{tabular}
\vspace{1mm}
\caption{Performances of trained and base model on test set data. These preliminary results point towards the separability of the accent axis in the corpus. The training procedure more than doubled the accuracy of the model. Also, the plot presented in Fig.~\ref{f:scatterplots_of_pca} show that the learned embedding tend to form clusters. Note that BN signifies batch normalization.}
    \label{t:quality_measures}. 
\end{table}

\section{Conclusion}
We presented SautiDB-Naija, an English database of non-native Nigerian accents to support the development of machine learning models for accent conversion or translation and classification tasks. We evaluated our corpus using mel-spectrogram and a fine-tuned wav2vec feature representations as input to an accent embedding model.
The results are promising, considering the small size of our corpus. Future work will focus on significantly expanding and diversifying the corpus while commencing research on L2 accent conversion tasks.

\section{Acknowledgement}
This manuscript is based in part on work supported by the AI4D-IndabaX Innovation Award, IDRC Grant Number: 109187-002, "SautiLearn : improving online learning experience with accent translation". Any opinions, findings and conclusions or recommendations expressed in this manuscript are those of the authors and do not necessarily reflect the views of IDRC. We would like to thank our colleagues Olumide Okubadejo and Munachiso Nwadike for their early contribution to this project.

\small
\bibliography{main}

\end{document}